\documentclass[sigconf]{acmart}

\AtBeginDocument{%
  \providecommand\BibTeX{{%
    \normalfont B\kern-0.5em{\scshape i\kern-0.25em b}\kern-0.8em\TeX}}}

\acmDOI{}
\acmISBN{}
\acmArticle{}
\acmPrice{}

\acmConference[recsysXfashion'19]
{Workshop on Recommender Systems in Fashion, 13th ACM Conference on Recommender Systems}
{September 20, 2019}
{Copenhagen, Denmark}
\acmYear{2019}
\copyrightyear{2019}

\begin{document}

\title[StyleRec]{Supporting stylists by recommending fashion style}


\author{Tobias Kuhn}
\thanks{Author Organization: Main contributor first, rest of authors organized alphabetically}
\author{Steven Bourke}
\author{Levin Brinkmann}
\affiliation{Outfittery GmbH}

\author{Tobias Buchwald}
\author{Conor Digan}
\author{Hendrik Hache}
\author{Sebastian Jaeger}
\affiliation{Outfittery GmbH}

\author{Patrick Lehmann}
\author{Oskar Maier}
\author{Stefan Matting}
\author{Yura Okulovsky}
\affiliation{Outfittery GmbH}

\renewcommand{\shortauthors}{Kuhn et al.}

\begin{abstract}
	Outfittery is an online personalized styling service targeted at men. We have hundreds of stylists who create thousands of bespoke outfits for our customers every day. A critical challenge faced by our stylists when creating these outfits is selecting an appropriate item of clothing that makes sense in the context of the outfit being created, otherwise known as style fit. Another significant challenge is knowing if the item is relevant to the customer based on their tastes, physical attributes and price sensitivity. 

At Outfittery we leverage machine learning extensively and combine it with human domain expertise to tackle these challenges. We do this by surfacing relevant items of clothing during the outfit building process based on what our stylist is doing and what the preferences of our customer are. In this paper we describe one way in which we help our stylists to tackle style fit for a particular item of clothing and its relevance to an outfit. A thorough qualitative and quantitative evaluation highlights the method's ability to recommend fashion items by style fit.

\end{abstract}


\maketitle

\section{Introduction}
\label{sec:intro}
One of the most important tasks in fashion recommendation is to answer questions like "Which shoes go well with this outfit?" Any solution has to capture compatibility between fashion items or, in other words, \textit{style fit}. But whether two fashion items fit together in style depends on many factors: low-level features such as color and texture, high-level features such as material and quality, and even less tangible features such as prominence and the association connected with the item. This challenge is particularly important for curated shopping services such as StitchFix, Zalon, or Thread - where every day thousands of stylists create personalized outfits for customers that must fit their stylistic needs. 

There are several publications dealing with the challenge of style fit. Veit et al.~\cite{veit2015} propose a siamese convolutional neural network to learn a transformation of fashion item images to a space representing compatibility between items. Items appearing together in a context are considered positive sample pairs when forming \textit{heterogeneous dyads}, i.e., belonging to different clothing categories. The thus learned space has the disadvantage that it doesn't account for the fact that style fit is not naturally a transitive property\footnote{In the case of a single style space, $\mathrm{fit}(A,B)\land \mathrm{fit}(B,C)\implies \mathrm{fit}(A,C)$, which is not necessarily true.}.
This shortcoming is addressed by Vasileva et al.~\cite{vasileva2018}, who propose to learn a separate style fit space for each pairing of categories. Furthermore, they incorporate accompanying textual descriptions to ensure semantic similarity.
Both of these works essentially deal with style fit between pairs. Han et al.~\cite{han2017} view outfit generation as related to sentences generation and hence propose a bi-direction long short-term memory network. This sequential approach allows their method to consider the whole outfit when suggesting a new item. Furthermore, Lee et al.~\cite{lee2017} apply two different convolutional neural networks to capture fashion semantics from outfit data by exploiting its images. 

All of these methods try to extract features from the fashion images and/or textual descriptions. But, as mentioned before, style fit depends on a variety of intangible features, some of which cannot be found in the considered input data - be it due to missing or wrong attributes or insufficient images. We therefore propose to learn a latent style embedding for each fashion item solely from the context in which they appear together by exploiting the curations and expertise of our in-house styling experts. 
To this end, we borrow from natural language processing: treating each item as a word and each outfit as context sentence, the popular word2vec~\cite{mikolov2013} method can be readily applied to learn each items location in a style fit space. Since a target and a context space are learned simultaneously, no transitive property of the style space is assumed. And, by allowing only heterogeneous dyads as sample pairs, the necessary inter-category compatibility is learned without confounding intra-category relations. Finally, we work on a granularity of  \textit{functional slots} rather than item categories to achieve a more natural clustering of items according to their function inside an outfit.
We then investigate different approaches to extend the trained pair style fit model to a proper outfit model, allowing for multi-item relations\footnote{Since $\mathrm{fit}(A,B)$ and $\mathrm{fit}(A,C)$ does not necessarily mean that $\mathrm{fit}(A,B+C)$.}.

A thorough quantitative and qualitative evaluation on our in-house dataset reveals the strengths and shortcomings of the proposed method. In the final section we discuss the possible implications and applications.

\section{Data}
\label{sec:data}
In this work, we define a fashion item without size information as a fashion \textit{product}. Any number of these products can be combined to form an \textit{outfit}, i.e., a set of products that can be worn together at the same time and fit together in fashion style. Every product is assigned a \textit{functional slot}, i.e., the role they fulfill in an outfit. A product can only ever fit in a single slot and an outfit can only ever be formed of products belonging to distinct slots. The full list of slots defined are
\begin{equation}
\begin{split}
\mathrm{slots} = \{ & \mathrm{shirt}, \mathrm{over\_shirt}, \mathrm{suit}, \mathrm{jacket}, \mathrm{belt}, \\
& \mathrm{trouser}, \mathrm{shoes}, \mathrm{other}\}.
\end{split}
\end{equation}

\begin{figure}[ht]
	\centering
	\includegraphics[width=\linewidth]{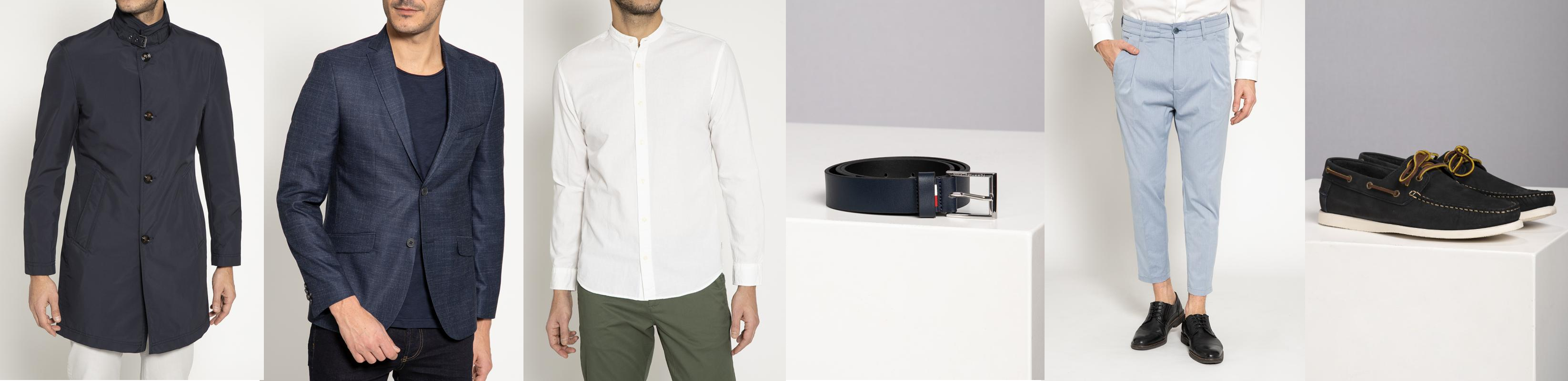}
	\caption{Typical outfit packed by a stylist at Outfittery. Functional slots from left to right are: \textit{jacket, suit, shirt, belt, trousers, shoes}.}
	\label{fig:outfit_example}
\end{figure}

As a curated shopping e-commerce platform, Outfittery employs fashion experts that compile outfits for the customers from a stock of products (see Fig.~\ref{fig:outfit_example} for an example). We assume that the products in these outfits fit together in style and hence use them as our ground truth.

Our dataset of outfits were generated as follows: (1) Take all outfits which have been sent out in the past. (2) Products that appear less than three times over the whole dataset are removed from the respective outfits. (3) If multiple products belonging to the same functional slot appear in one outfit, only one of them is randomly kept. (4) Remove outfits containing less than two products. We created a random sample of $\sim\!\!300,000$ outfits from this dataset, with $\sim\!\!6$ products on average drawn from $\sim\!\!20,000$ unique products. Furthermore, fashion products have a high turnover and our stock changes regularly. Hence, a model trained on the data from last year would not be applicable today. We therefore split the outfits into time windows, each containing $1,000$ outfits sent out consecutively. 

Train, validation, and test datasets are created by random uniform sampling over these time windows. The following subsections describe the sampling mechanisms used in order to create these datesets which are consumed by our models described in Sec.~\ref{sec:pairmodel} and~\ref{sec:outfitmodel}.

\subsection{Pair sampler}
\label{subsec:pair_sampler}
The pair model defined in Sec.~\ref{sec:pairmodel} is trained on positive and negative style fit product pairs. 

\subsubsection{Positive samples}
Positive samples are sampled from a finite set of outfits $\mathcal{O} = {\{\mathcal{O}_k\}}$, where each outfit $\mathcal{O}_k$ is a finite set of products, $\mathcal{O}_k=\{\mathrm{p}_i^k\,|\,p_i^k \in \mathcal{P}\}$. $\mathcal{P}$ is the set of all products which have been packed at least once. A set of positive samples $\mathcal{S}_\mathrm{pos}$ is formed from each pair of products appearing together in any outfit~$\mathcal{O}_k$
\begin{equation}
\begin{split}
    \mathcal{S}_\mathrm{pos} = \Bigl\{\bigl(\mathrm{p}_i^k, \mathrm{p}_j^k\bigr) \, | & \, \forall k=1,\ldots,|\mathcal{O}|\land
    \mathcal{O}_k\in \mathcal{O} \\ & \land \mathrm{p}_i^k\in\mathcal{O}_k\land \mathrm{p}_j^k\in\mathcal{O}_k\land i\neq j \Bigr\}.
\end{split}
\end{equation}
For reasons of simplicity, we drop in the following the index $k$ from product representations, such as $\mathrm{p}_i$.

Since each product in an outfit belongs to a distinct functional slot, the elements of each pair $(\mathrm{p}_i, \mathrm{p}_j)$ never share slots and hence form heterogeneous dyads. We define $\mathrm{p}_i$ as \textit{target} and $\mathrm{p}_j$ as \textit{context product}. Note, that the same pair can appear multiple times in the multiset $\mathcal{S}_\mathrm{pos}$, representing their frequency of appearance in the outfits. 

\subsubsection{Negative samples}
Since our dataset contains only positive pairs, we sample negative pairs from a background distribution using the negative sampling scheme~\cite{goldberg2014}. To obtain the negative samples, we hold the target product $\mathrm{p}_i$ of a positive sample $(\text{p}_i, \text{p}_j)$ and randomly draw $N_\mathrm{pair}$ negative samples $(\text{p}_i, \text{p}_n)$, with $n=1,\ldots,N_\mathrm{pair}$, where the negative context product $\text{p}_n$ is required to share the functional slot with $\text{p}_j$ and time window. This way, when we look for instance at a trouser-to-shoe relation, we train the model against the background noise of all trouser-to-shoe relations and not the other unrelated slot combinations. Furthermore, by sampling negative and positive pairs from the same time window, we ensure that both can be considered as drawn from distributions with the same support, i.e., as pairs of items from the available stock of that particular day. Otherwise, many negative pairs would consist of articles from different seasons and would, irrespectively of their style match, have no chance of occurring as positive pairs. See Sec.~\ref{sec:pairmodel} for more details on how we incorporate negative samples in the pair model definition.

\subsubsection{Subsampling}
Negative samples are drawn uniformly over all products from the same time window as the positive samples, thereby representing the underlying frequency distribution: products that are seen more often in outfits are picked more often.
Taking the same approach for positive samples would hurt the representation of less frequent products. Hence, we employ the subsampling strategy proposed in Mikolov et al.~\cite{mikolov2013} and discard each positive context product $\text{p}$ in an outfit with the probability of
\begin{equation}
    p(discard|\mathrm{p}) = 1 - \max\left(\sqrt{\rho / f(\mathrm{p})}, 0\right)\!,
\end{equation}
where $f(\mathrm{p})$ is the frequency of appearance of product $\mathrm{p}$ in the outfits of a time period and $\rho$ an empirically determined threshold parameter. Effectively that means that products which appear with a frequency lower than $\rho$ are more likely to be picked as positive sample than dictated by their frequency.

\subsection{Outfit sampler}
\label{subsec:outfit_sampler}

Samples for outfit model training and evaluation are generated from the dataset defined in Sec.~\ref{sec:data}. To obtain a balanced dataset, we sample from each outfit $\mathcal{O}_k$ subsets of products of size $1,...,|\mathcal{O}_k|-1$. For each subset we consider one of the remaining products of the original outfit $\mathcal{O}_k$ as query product. To each pair of query product and subset, $N_\mathrm{outfit}$ negative samples are generated out of the same functional slot and the same time window.

\section{Pair model}
\label{sec:pairmodel}
We define a \textit{pair model} as a function that takes a reference product, $\mathrm{p}_\mathrm{ref}$, and a query product, $\mathrm{p}_\mathrm{query}$, and returns a numeric score that reflects their style fit
\begin{equation}
f_{\mathrm{p}}\colon(\mathrm{p}_\mathrm{ref}, \mathrm{p}_\mathrm{query}) \mapsto \{x\in \mathbb{R}| -1 \leq x \leq 1\}.
\label{eq:func}
\end{equation}

\subsection{word2vec based model}

Since its introduction a few years back~\cite{mikolov2013}, the  embedding technique word2vec has become a wide spread concept in the machine learning community. It is a small neural network that learns embeddings for each word of a corpus. By computing the cosine similarity between two embeddings for distinct words, a measure of their transitional properties is obtained: \textit{king} and \textit{crown} might for example have a higher likelihood of co-occurrence than \textit{accountant} and \textit{crown}. For negative sampling random pairs from all words are drawn, known as negative sampling~\cite{goldberg2014} or noise sampling. Training this model on all of these samples, tries to assign high probabilities to real context words and low probabilities to noise context words. The idea of word2vec has been successfully extended to other entities: code~\cite{alon2019}, genes~\cite{du2019}, or, more generally, item2vec~\cite{barkan2016}.

We propose to employ this method to learn suitable vector representations for products that represent how well they fit together in style. In analogy to the word2vec concept, the products are words and the outfits are sentences. For each product in an outfit, all other products in the same outfit are considered as context. Following Sec.~\ref{sec:intro} unique product identifiers are used to represent a word, while additional contextual information such as image data or attributes are not taken into account. 

Formally, the model is defined as follows: Given a finite set of products $\mathrm{p}_i\in \mathcal{P}$ and a number of context sets $\mathcal{O}_k\in \mathcal{O}$ with size $|\mathcal{O}_k| \leq |\mathrm{slots}|$, the model aims to maximize the average conditional $\mathrm{log}$ 
probability 
\begin{equation}
    \sum_{k=1}^{|O|}\frac{1}{|\mathcal{O}_k|}\sum_{i=1}^{|\mathcal{O}_k|}\sum_{j\neq i}\log p(\text{p}_i|\,\mathrm{p}_j)
    \label{fig:lion}.
\end{equation}
Following the idea of negative sampling~\cite{goldberg2014}, $p(\mathrm{p}_i|\,\mathrm{p}_j)$ is defined as
\begin{equation}
    p(\text{p}_i|\mathrm{p}_j) = \sigma\!\left(u_i^T v_j\right)\prod_{l=1}^{N_\mathrm{pair}}\sigma\!\left(-u_i^Tv_l\right)
    \label{eq:pairmodel}\!,
\end{equation}
where $u_i\in U(\subset\varmathbb{R}^m)$ and $v_i\in V(\subset\varmathbb{R}^m)$ are $m$-dimensional latent vectors representing the target and context of product $\mathrm{p}_i$, $N_\mathrm{pair}$ determines the number of negative samples per positive sample, and $\sigma(x) = 1/(1+\exp(-x))$. Both $N_\mathrm{pair}$ and $m$ are determined empirically.

\subsection{Style fit score}
In order to calculate a style fit score between two products $(\mathrm{p}_i, \mathrm{p}_j)$ the cosine similarities $\text{sim}(u_i, v_j)$ and $\text{sim}(u_j, v_i)$ across target and context space are averaged

\begin{equation}
    f_{p}(p_i, p_j) = \frac{1}{2} \left(\text{sim}(u_i, v_j) + \text{sim}(u_j, v_i)\right)\!. 
    \label{eq:prediction_pairmodel}
\end{equation}
Note that cosine similarities between the target vectors, $\text{sim}(u_i,u_j)$, and context vectors, $\text{sim}(v_i,v_j)$, simply express the similarity within the respective embedding spaces and thus similarity in style itself, but not their style fit to each other.

\section{Outfit models}
\label{sec:outfitmodel}
\label{sec:outfit_model}
The above proposed pair model can predict the style fit between two products. Another use case to be considered in the scope of this work is the completion of an \textit{incomplete outfit} $\mathcal{\tilde{O}}$. That means, the question is to find the best matching new product $\mathrm{p}_i$ to a set of fixed products, defined as $\mathcal{\tilde{O}}$. An \textit{outfit model} is defined as a function that takes an (incomplete) outfit $\mathcal{\tilde{O}}$, and a query article, $\mathrm{p}_\mathrm{query}$, and returns a numeric score that reflects their style fit 
\begin{equation}
f_\mathrm{O}\colon (\mathrm{p}_\mathrm{query}, \mathcal{\tilde{O}}) \mapsto \{x\in \mathbb{R}| -1 \leq x \leq 1\}.
\end{equation}
In this section we present two outfit models to approach this challenge.

\subsection{Mean model}
\label{subsec:mean_model}
Based on interviews with our stylists, we concluded that outfit composition might be reduced to a sum of pair interactions between the products within an outfit. Following the assumption of independence, we can therefore model the matching score of the new product $\text{p}_i$ to an outfit $\mathcal{\tilde{O}}$ as
\begin{equation}
    f_{\mathrm{O}}^\mathrm{M}(\mathrm{p}_i, \mathcal{\tilde{O}}) = \frac{\sum_{\mathrm{p}_j\in \mathcal{\tilde{O}}} f_\mathrm{p}(\mathrm{p}_i, \mathrm{p}_j)}{|\mathcal{\tilde{O}}|}
    \label{eq:meanmodel},
\end{equation}
where $f_\mathrm{p}(\mathrm{p}_i, \mathrm{p}_j)$ is the response of the pair model as defined in Eq.~\ref{eq:prediction_pairmodel}. Since this mean model is parameter free, it requires no training beyond the underlying pair model.

\subsection{Attention model}
It is reasonable to assume that the combination of the functional slots of the reference product $\mathrm{p}_i$ and the new query product $\mathrm{p}_j$ has an impact on how strong the associated pair model score should be weighted. For example: the choice of a belt might highly depend on the selected shoes and less on the selected jacket. To account for this, we reformulate Eq.~\ref{eq:meanmodel} to

\begin{equation}
    f_\mathrm{O}^\mathrm{A}(\mathrm{p}_i, \mathcal{\tilde{O}}) = \sum_{\mathrm{p}_j\in \mathcal{\tilde{O}}}\alpha_{s_i,s_j} f_{p}(\mathrm{p}_i, \mathrm{p}_j)
    \label{eq:attentionmodel},
\end{equation}
where $\alpha_{s_i,s_j}$ is a trainable parameter depending on the functional slots $s_i$ and $s_j$ of the two products $p_i$ and $p_j$, respectively. Note it is asymmetry, i.e., the impact from shoes to shirts, $\alpha_{s_{\text{shoe}}, s_{\text{shirt}}}$, might differ from the impact from shirts to shoes, $\alpha_{s_{\text{shirt}}, s_{\text{shoe}}}$. The implementation is realized as a neural network based on a simplified attention mechanism by Vaswani et al.~\cite{vaswani2017} followed by a soft-max layer to ensure that $\sum_{\mathrm{p}_j\in \mathcal{\tilde{O}}}\alpha_{s_i,s_j} = 1$. This set-up allows the attention model to weight each functional slot pairing differently.

\section{Experiments}
In this section, we present the experimental set-up and we provide qualitative insights for the described pair and outfit models. In conclusion the results are compared with Vasileva et al.'s  method ~\cite{vasileva2018}. 

\subsection{Experimental set-up}

\subsubsection{Pair Model}
\label{subsubsec:pair_model_exp_setup}

The model training applies AdaGrad (Duchi et al. ~\cite{duchi2011}) - an adaptive gradient descent method - with a learning rate of 1.0. The optimization runs 30 epochs over the training set. Following Sec.~\ref{subsec:pair_sampler}, we add $N_\mathrm{pair}=80$ negative sampled pairs to each positive heterogeneous dyad. The positive sample discarder parameter $\rho$ is set to 0.0002. 

To evaluate the performance of the models we create test and train splits, where each test and train instance is a set of products consisting of 1 positive and 19 negative samples. For each instance we do listwise evaluation, where we compute precision at 2, reported as Top 2 score. We use the equation Eq.~\ref{eq:prediction_pairmodel} to compute each permutation. 


For illustrative purposes, we also report on hit rate for each rank position in our list (1 - 20) such that we can demonstrate where in the list the majority of positive samples are placed. In this case we consider the hit rate as $1 / \mathrm{rank}$.

\subsubsection{Outfit Model}

The mean model uses the trained pair model as a base and requires no further training. The attention model is using the same optimization framework as stated in Sec.~\ref{subsubsec:pair_model_exp_setup} and is trained over 10 epochs. According to Sec.~\ref{subsec:outfit_sampler}, $N_\mathrm{outfit}=19$ negative query products are added to each pairing sample outfit and positive query product pair for training. 

We evaluate the model using the hit rate at the ranked position, Average Precision Score ($\mathrm{APS}$) and the Fill-in-the-Blank ($\mathrm{FITB}$) accuracy. The $\mathrm{APS}$ summarizes the precision-recall curve as the weighted mean of precisions achieved at each threshold~\cite{zhu2004}. For the $\mathrm{FITB}_n$ accuracy, one randomly selected product of an outfit of size $n$ is kept fix together with $n-1$ randomly sampled products sharing the same functional slot. The goal is to select the positive product as ranked highest. 

Additionally, we baseline these models against the work by Vasileva et al.~\cite{vasileva2018}. We transformed our dataset with the same train-test-split as our other experiments into the Polyvore format and evaluated their pretrained model\footnote{\url{https://github.com/mvasil/fashion-compatibility}} on our test set. The comparison is based on the $\mathrm{FITB}_4$ score.

\section{Results}
In this section we describe the results of our different experimental evaluations. 

\subsection{Visualization}
\begin{figure}
    \centering
    \includegraphics[width=\linewidth]{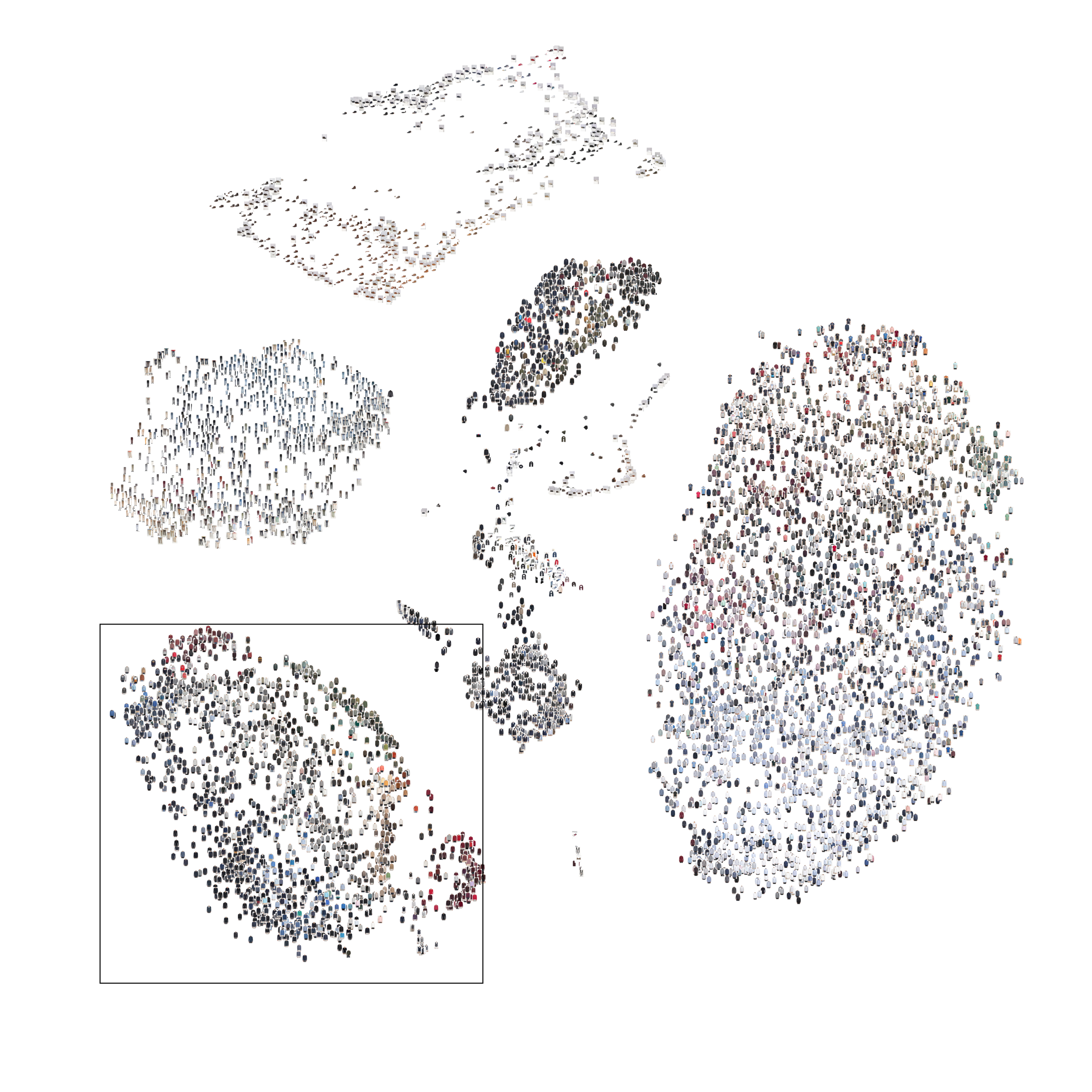}
    \caption{t-SNE plot of the pair model item embedding within the target space. The area at the left bottom is shown in Fig.~\ref{fig:tsne_pairmodel_zoomed}.}
    \label{fig:tsne_pairmodel}
\end{figure}

\begin{figure}
    \centering    
    \includegraphics[width=\linewidth]{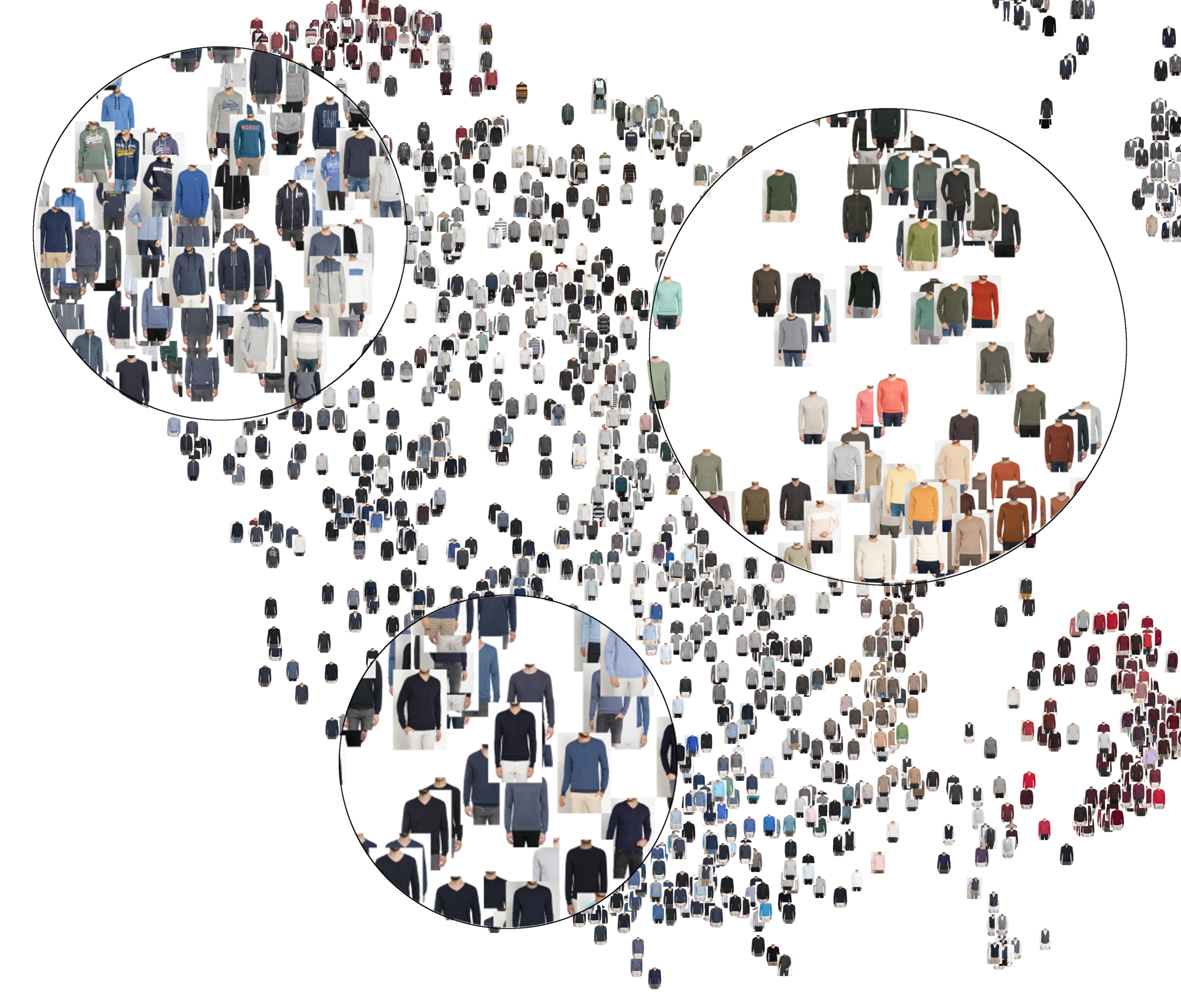}
    \caption{Detailed view on overshirt area of the t-SNE plot in Fig.~\ref{fig:tsne_pairmodel}.}
    \label{fig:tsne_pairmodel_zoomed}
\end{figure}

We visualize the embedding of our products into target space with t-SNE [7]. Fig.~\ref{fig:tsne_pairmodel} shows a high-level clustering into functional slots. Within these slots we see clustering of items by the most important stylistic features, such as patterns, color, item type, and gradual changes from formal to casual. Fig.~\ref{fig:tsne_pairmodel_zoomed} visualizes these stylstic differences within the functional slot for overshirts. Both figures
support the finding that our embeddings capture important stylistic
features of a product.

\subsection{Pair Model}
\begin{figure}
    \centering
    \includegraphics[width=\linewidth]{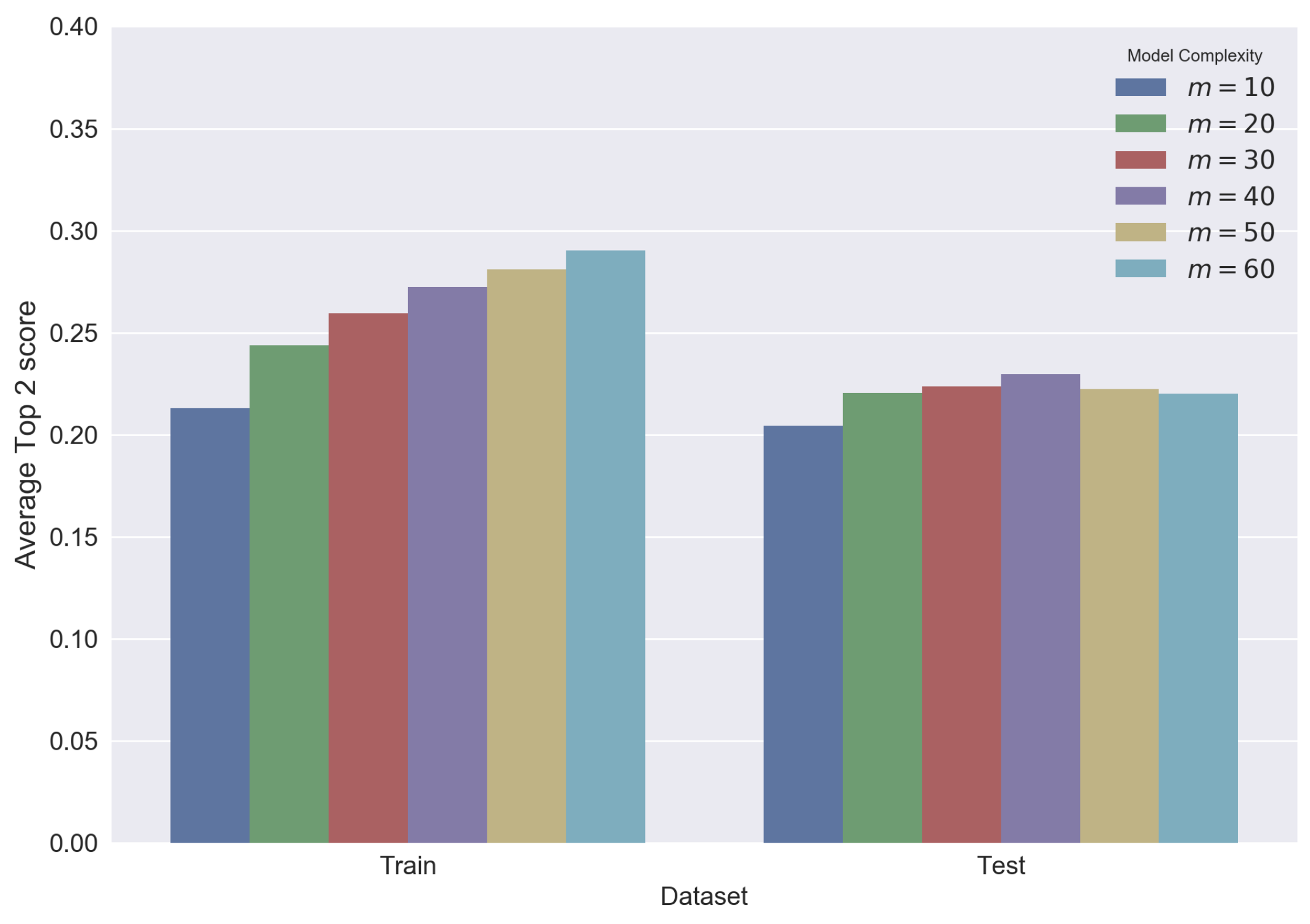}
    \caption{Top 2 score for varying model complexity parameter $m$ values.}
    \label{fig:pair_top2}
\end{figure}

Fig.~\ref{fig:pair_top2} displays the Top 2 score on the test and the train set for various values of the model complexity parameter $m$. Increasing the model complexity improves the evaluation performance on the train set. This appears to be an indication of overfitting. As can be seen in the chart, the best performing parameter gets a value of 0.28.  We can see that in the test split the overall best performing value for $m$ is at 40, which has a value of 0.23.

\begin{figure}
    \centering
    \includegraphics[width=\linewidth]{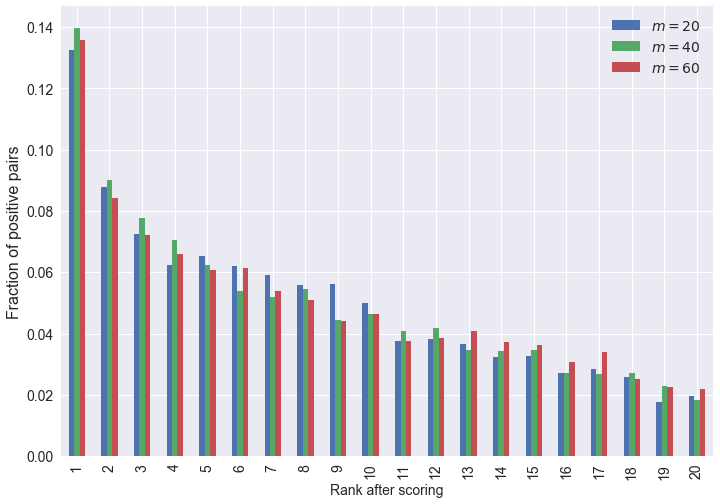}
    \caption{Averaged Hit Rate}
    \label{fig:pair_fraction_over_rank}
\end{figure}

Fig.~\ref{fig:pair_fraction_over_rank} shows the averaged hit rate at different positions in the list. We observe that $m=40$ performs best up until position 4.  

\subsection{Outfit Model}

\begin{figure}
    \centering
    \includegraphics[width=\linewidth]{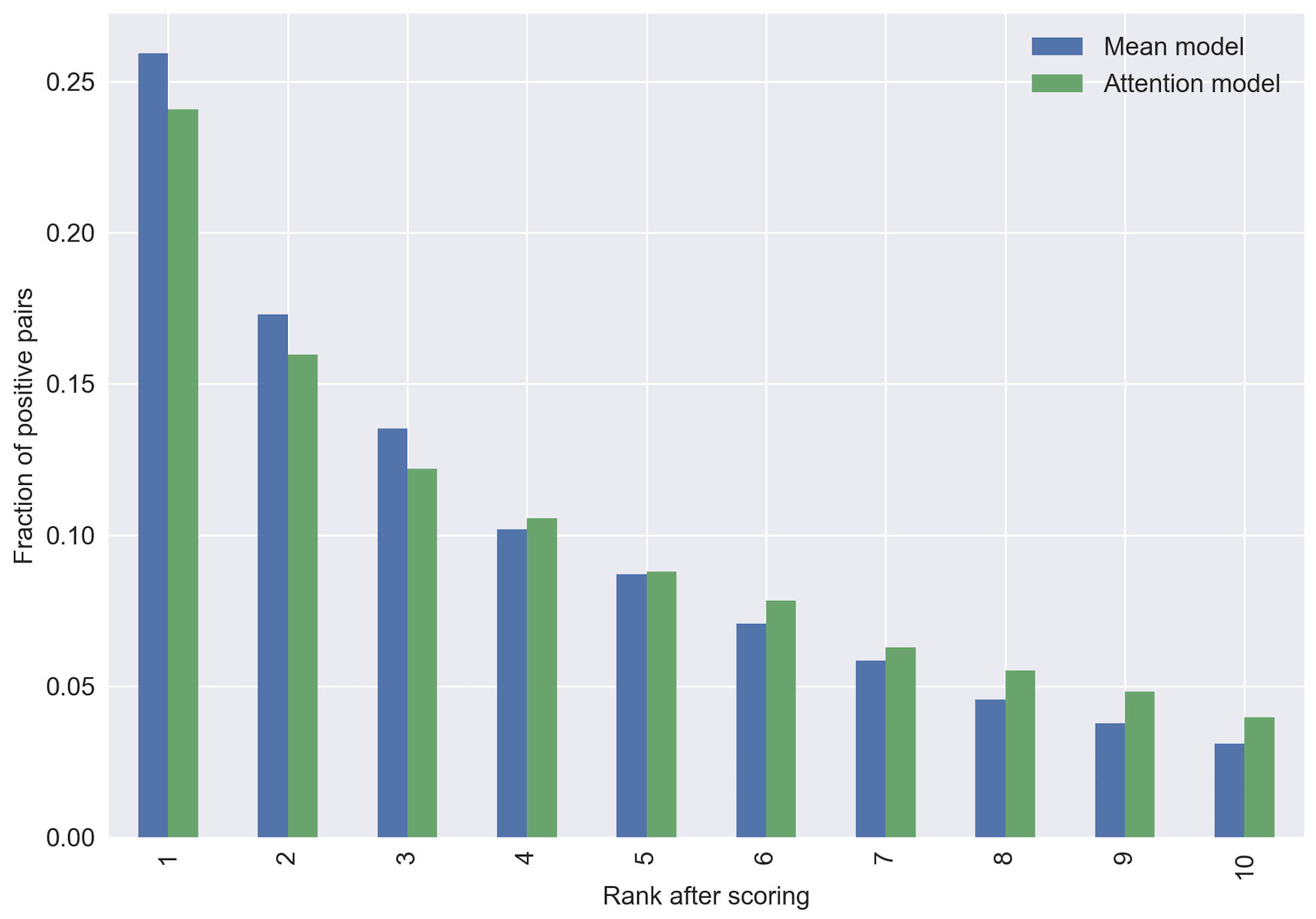}
    \caption{Hit rate at the ranked position for outfit models.}

    \label{fig:outfit_fraction_over_rank}
\end{figure}

Fig.~\ref{fig:outfit_fraction_over_rank} shows the hit rate at different ranks. The mean model assigns higher values for top scores and lower values for bottom scores compared to the attention model. 

In Table~\ref{tab:outfit_metrics} the performance of our approaches against the Vasileva model is shown. Firstly, it is worth noting that Vasileva's work is not 100\% comparable here due to a variety of differences in the underlying data. Vaisleva's model performs substantially worse on our data than in their data ($\mathrm{FITB}_4$ of 0.317 vs 0.576)~\cite{vasileva2018}. None the less, we believe it to be a reasonable baseline for the task at hand. What we can see from the results is in relation to $\mathrm{FITB}$ based metrics the mean model outperforms the other approaches. 

\begin{table}
    \caption{Comparison of outfit models for the Fill-in-the-blank metrics $\mathrm{FITB}_{10}$ and $\mathrm{FITB}_{4}$ and the Average Precision Score (APS) applied on Outfittery's dataset.}
    \label{tab:outfit_metrics}
    \begin{tabular}{llccc}
        \toprule
        Model & Dataset & $\text{FITB}_{10}$ & $\text{FITB}_{4}$ & $\text{APS}$ \\
        \midrule
        Mean Model & Outfittery & 0.258 & 0.471 & 0.366 \\
        Attention Model & Outfittery & 0.239 & 0.442 & 0.342 \\
        Vasileva's model & Outfittery & &  0.317 &  \\
        \bottomrule
    \end{tabular}
\end{table}

\section{Applications}
\begin{figure}
    \centering
    \includegraphics[width=\linewidth]{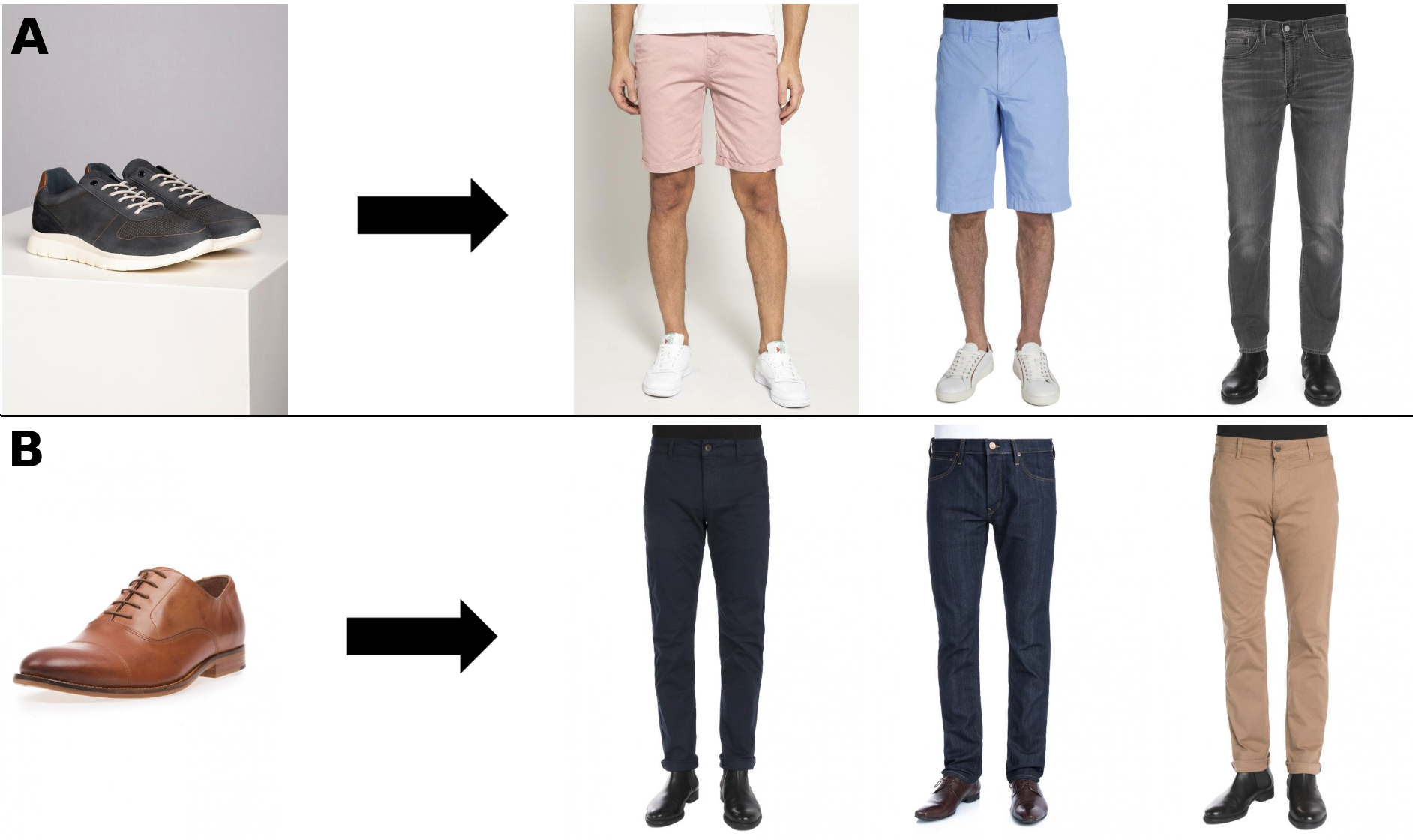}
    \caption{Example of stock ranking by style fit using the pair model. (A) Using a casual shoe and (B) a business shoe as reference article, respectively.}
    \label{fig:ranking3}
\end{figure}

In this section we describe a few ways in which we exploit our models in our various different systems at Outfittery. In general we find this approach to be but one of many useful ways to help discover and improve style fit. We can successfully use this for article ranking. For example the pair model allows us to sort our stock by style fit to a given article. Fig.~\ref{fig:ranking3} show the top three ranked articles given two distinct reference pairs of shoes. The ranking can be naturally extended using the mean model.

Furthermore, one application of having an outfit model as described in Sec.\ \ref{sec:outfit_model} is automated outfit creation. We propose to use our mean model for automated outfit composition by the following beam search \cite{beamsearch} procedure that is also used in sequence-to-sequence language generation tasks.

We define a fixed order of the functional outfit slots and a beam width $b$. For the first slot we select $b$ random products as starting outfits. Then we continue adding candidate products for the next slot to each outfit. The resulting outfits are scored using the mean model, keeping only the top $b$ outfits in each step. 

\begin{figure}
    \centering
    \includegraphics[width=\linewidth]{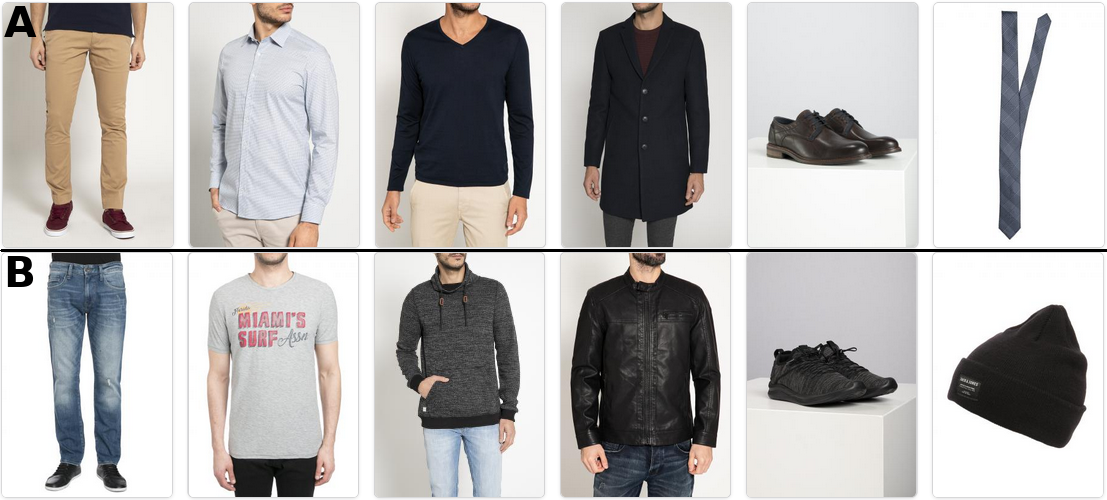}
    \caption{Automatic generated outfit with (A) beam width 1 and (B) beam width 20.}
    \label{fig:outfit_bs}
\end{figure}

The qualitative results (see Fig.~\ref{fig:outfit_bs}) look compelling. We are aware that this model tends to prefer popular products. It remains to investigate if such outfits are diverse enough.

\section{Discussion}
The presented qualitative and quantitative results show that item embeddings in latent space allow to tackle the question of style fit both for item-to-item and also item-to-outfit relations.

The experiments on the outfit model suggest that the mean model outperforms the more complex attention model. A possible explanation is that the combinations of categories is already incorporated in the pair score implicitly. 

The comparison to Vasileva's method reveals that simply re-using their model with our stock images hardly outperforms random scoring. This can be explained by a very different type of data, i.e., women fashion images and text attributes, used for training their model. Even though the datasets and models and thus the $\mathrm{FITB}$ scores are not fully comparable, the $\mathrm{FITB}_{4}$ accuracy of Vasileva's model on their data compared to our model on our data ($\mathrm{FITB}_4$ 0.567 vs 0.471) indicates room for improvement by exploiting additional features, such as image or attribute data. 

One potential limitation of our proposed approach is the cold start problem. However as this work is currently used daily by hundreds of stylists this is not a practical concern as we get new training data every day.


Using only curated outfits to infer embeddings allows Outfittery to tackle the question of style fit without the usage of additional attribute or image data. Furthermore, the described models are deployed in production and strongly support our stylist teams in their daily work. 

\subsection{Acknowledgements}
We would like to thank the reviewers for their thoughtful comments and feedback. We would also like to thank our internal data platform and infrastructure teams for making this work possible by providing outstanding tools and support. 

\bibliographystyle{ACM-Reference-Format}
\bibliography{bibliography}

\end{document}